%% file: main.tex
\documentclass[10pt,twocolumn,letterpaper]{article}

\usepackage{wacv}              % To produce the CAMERA-READY version

\usepackage[subtle]{savetrees} % A package designed to gently pack text

\definecolor{wacvblue}{rgb}{0.21,0.49,0.74}
\usepackage[pagebackref,breaklinks,colorlinks,allcolors=wacvblue]{hyperref}

\def\wacvPaperID{5} % *** Enter the WACV Paper ID here
\def\confName{CV4EO}
\def\confYear{2026}

\title{DIS2: \textit{Dis}entanglement Meets \textit{Dis}tillation with Classwise Attention for Robust Remote Sensing Segmentation under Missing Modalities}

\author{Nhi~Kieu\textsuperscript{1} \and Kien~Nguyen\textsuperscript{1} \and Arnold~Wiliem\textsuperscript{2} \and Clinton~Fookes\textsuperscript{1} \and Sridha~Sridharan\textsuperscript{1}\\
\textsuperscript{1}Queensland University of Technology, Australia \qquad \textsuperscript{2}Shield AI, Australia\\
{\tt\small \{v.kieu,k.nguyenthanh,c.fookes,s.sridharan\}@qut.edu.au}\\
{\tt\small arnold.wiliem@shield.ai}
}
\begin{document}
\maketitle
\input{sec/0_abstract}

\input{sec/1_intro}
\input{sec/2_related_work}
\input{sec/3_method}
\input{sec/4_experiment}
\input{sec/5_conclusion}
{
    \small
    \bibliographystyle{ieeenat_fullname}
    \bibliography{main}
}
\end{document}

%% file: sec/0_abstract.tex
\begin{abstract}
The efficacy of multimodal learning in remote sensing (RS) is severely undermined by missing modalities. The challenge is exacerbated by the RS highly heterogeneous data and huge scale variation. Consequently, paradigms proven effective in other domains often fail when confronted with these unique data characteristics. Conventional disentanglement learning, which relies on significant feature overlap between modalities (modality-invariant), is insufficient for this heterogeneity. Similarly, knowledge distillation becomes an ill-posed mimicry task where a student fails to focus on the necessary compensatory knowledge, leaving the semantic gap unaddressed. Our work is therefore built upon three pillars uniquely designed for RS: (1) principled missing information compensation, (2) class-specific modality contribution, and (3) multi-resolution feature importance. We propose a novel method \textbf{DIS2}, a new paradigm shifting from modality-shared feature dependence and untargeted imitation to \textbf{active, guided missing features compensation}. Its core novelty lies in a reformulated synergy between disentanglement learning and knowledge distillation, termed DLKD. Compensatory features are explicitly captured which, when fused with the features of the available modality, approximate the ideal fused representation of the full-modality case. To address the class-specific challenge, our Classwise Feature Learning Module (CFLM) adaptively learn discriminative evidence for each target depending on signal availability. Both DLKD and CFLM are supported by a hierarchical hybrid fusion (HF) structure using features across resolutions to strengthen prediction. Extensive experiments validate that our proposed approach significantly outperforms state-of-the-art methods across benchmarks. The source code is made {available}\footnote{https://github.com/nhikieu/DIS2}.
\end{abstract}

%% file: sec/1_intro.tex
\section{Introduction}
Multimodal learning has become a cornerstone of computer vision in various domains \cite{zhang2020advances, zhang2021deep}. However, heterogeneous signals in RS pose unique challenges, particularly in scenarios with missing modality \cite{kieu2024multimodal}. For example, spectral data from RGIR (Red-Green-Infrared) carry material and appearance cues, whereas NDSM (Normalized Digital Surface Model) encodes structural height information. Leveraging both modalities yields a significant performance boost, but the absence of one leads to collapsed predictions. Real-world deployments of RS applications are susceptible to incomplete multimodal signals due to weather interference and sensor malfunction \cite{kieu2024multimodal}. Improving the model reliability under missing modalities has gained attention in many domains such as Medical Imaging \cite{zhou2023literature, zhang2020advances} and Vision-Language-Action (VLA) \cite{wu2024deep}. However, for fine-grained semantic segmentation in RS, enhancing robustness towards missing modalities remains underexplored.

Existing strategies can be categorized into four groups: generative models, disentanglement learning, knowledge distillation and other implicit methods that have efficiency as their competitive edge but at the cost of performance trade-off. They are limited in addressing the core problem of missing modality: information compensation. Generative approaches \cite{ma2021smil, dalmaz2022resvit, zhou2020hi} attempt to explicitly reconstruct the original missing signals. While powerful, they often fall short in capturing sophisticated synergies between heterogeneous inputs due to limitations of early layer raw signals restoration. On the other hand, conventional disentanglement learning architectures \cite{yao2024drfuse, wang2023multi} decompose features into modality-shared and modality-specific subspaces. While promising, they provide a passive decomposition of features lacking an active mechanism to reconstruct the semantic contribution of the absent modality. Knowledge distillation methods \cite{dai2024study, du2024tip} blindly imitate full-modality teacher representations instead of learning the specific knowledge needed to compensate for the loss from the missing modality while leveraging the student's own capability.

We argue that \textit{a truly robust solution requires a paradigm shift from pixel-level reconstruction, passive decomposition, and untargeted imitation to active and guided feature compensation} in the latent space. Thus, we introduce \textbf{DIS2}, a novel framework built on this principle and uniquely suited to the challenges of RS where discriminative cues could reside predominantly in a single modality and small, rare classes are overlooked. To capture varying contributions of each modality by target class and by scenarios, our design consists of 3 synergistic components:
\begin{itemize}
    \item DLKD: The \textit{reformulated disentanglement} is designed to provide non-redundant distinct and supplementary feature spaces to actively learn compensatory representations. The structured compensation process is guided by \textit{hierarchical distillation} loss, which transfers privileged knowledge from the full modality setting to missing modality scenarios. Adaptive branch activation allows the model to handle arbitrary modality combinations.
    \item Hierarchical Hybrid Fusion (HF): progressively fuse multimodal features through convolution and transformer. It fosters the capturing of complex interactions between modalities and supports DLKD/CFLM across scales.
    \item Classwise Feature Learning Module (CFLM): introduces (i) classwise attention maps generated based on learnable class-relevant queries and (ii) a classwise multiscale decoder. Together, they enable selective feature attention for each class, improving recognition under severe class imbalance and huge scale variation.
\end{itemize}

\vspace{3px}
\noindent The contributions of this study are threefold:
\begin{itemize}
    \item We propose \textbf{DIS2} - a novel unified architecture uniquely tailored to RS characteristics as described above.
    \item Extensive experiments are conducted on two popular semantic segmentation RS datasets - Vaihingen, Potsdam - demonstrate the superiority of our method over representative SOTA models across families including GEMMNet \cite{kieu2025gemmnet}, mmformer \cite{zhang2022mmformer}, ShaSpec \cite{wang2023multi}, SimSiam \cite{chen2021exploring} and its variant SimSiamTransfer.
    \item The source code and trained weights will be made publicly available to foster transparency and accelerate future research in this direction.  
\end{itemize}

%% file: sec/2_related_work.tex
\section{Related Works}
\textbf{Multimodal Learning.} The fusion of heterogeneous data is essential to remote sensing, as the combination of spectral information (e.g., RGIR) and structural data (e.g., NDSM) provides a far richer scene representation than any single sensor can offer \cite{kieu2024multimodal, zhang2021deep}. Fusion strategies have evolved from simple concatenation and advanced fusion mechanism in early or late stages of a network \cite{kieu2023general, zhang2024multimodal, fan2023pmr} to more balanced, sophisticated intermediate fusion mechanisms \cite{hong2020more, li2022dense, zhang2020advances}. Beyond the fusion stage, the mechanism for combining features is equally critical. A hybrid architecture, that leverage both convolution for their strength in extracting local patterns and transformer for their ability to model global context, emerge as a leading approach \cite{dai2021transmed, zhang2022mmformer, kieu2025gemmnet}. On the other hand, significant scale variation inherent in remote sensing data can be mitigated through multiscale learning \cite{kieu2023general}, which has also been proven to be beneficial for fine-grained semantic segmentation task \cite{tang2022matr, zhou2020hi}. Despite these advancements, existing architectures are not explicitly designed to be robust against the dynamic absence of an entire input stream, which can lead to feature space collapse or instability. \textit{Our Hierarchical Hybrid Fusion (HF) structure builds on these established principles of hybrid, multiscale fusion. Moreover, it is uniquely architected to provide rich feature foundations for our other modules (DLKD and CFLM) under any scenario.}

\vspace{3px}
\noindent\textbf{Missing Modality.} 
Handling missing modalities is a critical hurdle in multimodal learning, attracting attention in high-stakes domains like medical imaging \cite{zhou2023literature, zhang2020advances}, VLA \cite{zhu2024review} and autonomous systems \cite{wu2024deep}. While existing paradigms have demonstrated promising performance, they address the problem from disparate angles. They can be categorized into four main groups: generative models, disentanglement learning, knowledge distillation and other implicit methods. Generative models such as hallucination networks \cite{wu2024deep}, Generative Adversarial Networks (GANs) \cite{dalmaz2022resvit, zhou2020hi} and Variational Autoencoders (VAEs) \cite{daunhawer2021limitations, ma2021smil} are among the most intuitive ones, attempting to explicitly reconstruct missing signals. However, this is an ill-posed task, particularly for heterogeneous data where the mapping between modalities is complex and non-linear \cite{yao2024drfuse, daunhawer2021limitations}. Disentanglement learning approaches \cite{yao2024drfuse, wang2023multi} separate shared and modality-specific features but ignore the active restoration of transferable privileged features. On the other hand, Knowledge Distillation methods \cite{chen2021exploring, du2024tip, dai2024study} often restrict distillation to logit-level untargeted mimicry. \textit{Our architecture reformulates disentanglement learning and knowledge distillation into a synergistic strategy called DLKD. It avoids redundancy while forcing the model to learn multi-level compensatory features that, when combined with discriminative modality-specific features, approach full-modality performance under missing modality conditions.}

\vspace{3px}
\noindent\textbf{Missing Modality in Remote Sensing}.
Despite advancements in methods in other domains for dealing with missing modality, the challenge in RS remains underexplored, especially for fine-grained prediction tasks like semantic segmentation \cite{kieu2025gemmnet}. Early and influential works focused primarily on generative approaches to ``hallucinate'' or reconstruct the missing data stream \cite{kampffmeyer2018, kumar2021improved, kieu2025gemmnet}. Their limitation is their ill-posed task of focusing on pixel-level fidelity rather than compensating for the semantic gap under a missing modality. A few more recent approaches have shifted towards disentanglement learning \cite{li2022dense} and knowledge distillation \cite{kang2022disoptnet, wei2023msh}. However, most existing studies are land cover classification \cite{wei2023msh} and binary building footprint segmentation \cite{chen2024novel} rather than fine-grained semantic segmentation. On the other hand, they overlook the reality that the importance of a modality is highly dependent on the semantic class and modality availability. This creates a significant research gap for a holistic approach that unifies and leverages existing advancements' respective strengths. \textit{The proposed method aims to address three critical gaps: (i) lack of robust hierarchical fusion that can foster the learning of (ii) compensatory information and (iii) class-aware modeling. Our \textbf{DIS2} is designed with three synergistic components \textbf{DLKD}, \textbf{HF} and \textbf{CFLM} to address these gaps.} The results demonstrate improved robustness under missing modality scenarios, especially for underrepresented classes.

%% file: sec/3_method.tex
\section{Methodology}
\label{sec:method}
This section presents the details of our \textbf{DIS2} framework, which integrates three synergistic components uniquely designed to tackle the challenge of RS data under missing modality scenarios: (1) DLKD \Cref{subsection:dlkd} - a reformulated synergy of disentanglement learning and knowledge distillation to explicitly capture compensatory features that once fused with modality-distinct features approaching the ideal full-modality prediction, (2) HF \Cref{subsection:hf} - a multi-resolution feature fusion structure facilitating the other two components and (3) CFLM \Cref{subsection:cflm} - a class-aware feature learning module with adaptability to different missing modality cases.

\begin{figure*}[t]
  \centering
  \includegraphics[width=0.90\textwidth]{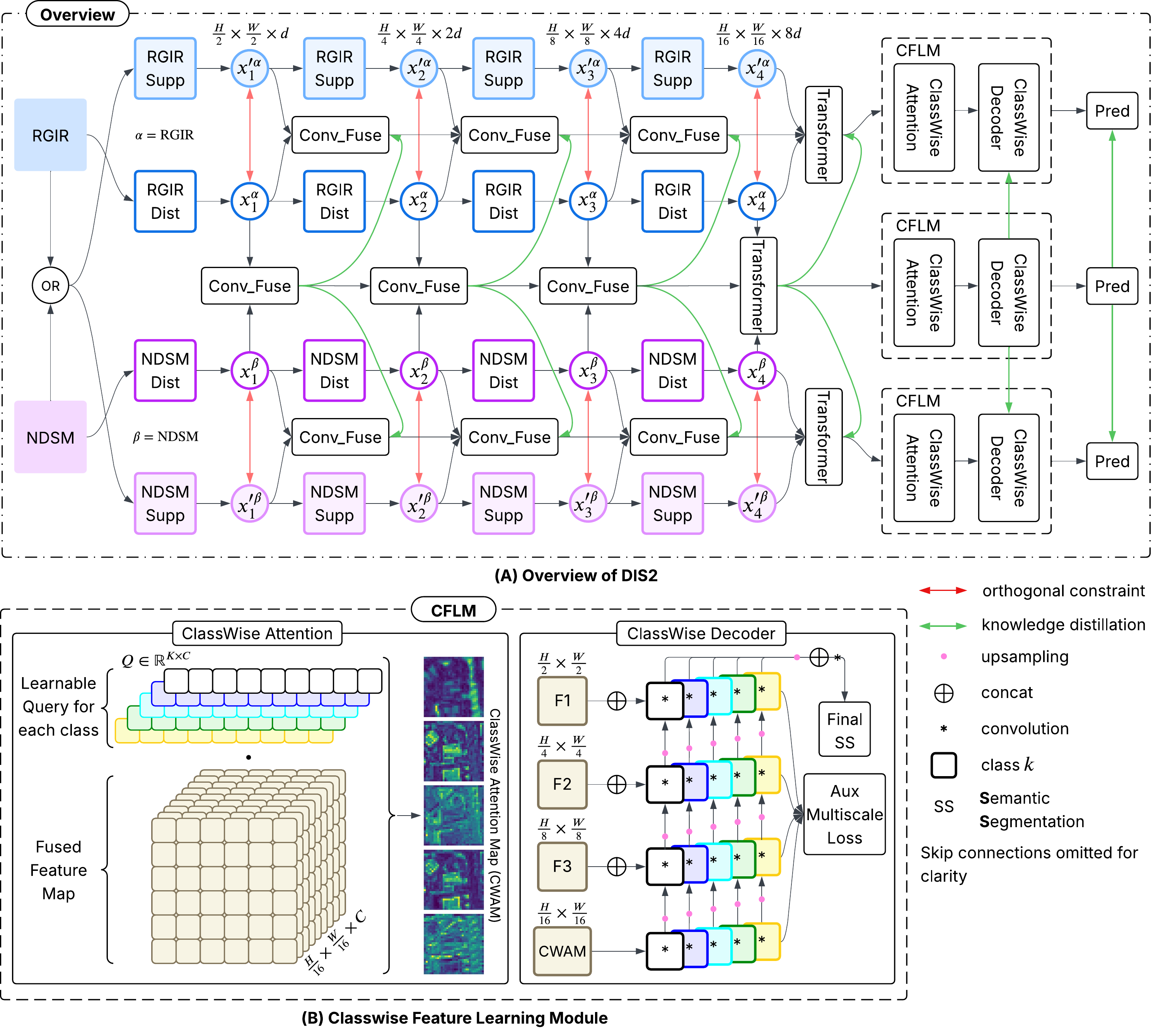}
  \caption{Illustration of our proposed method}
  \label{fig:model_schema}
\end{figure*}

\noindent \Cref{fig:model_schema}A illustrates the high-level workflow of our architecture. Input images come from different modalities $\boldsymbol{I}^m$ where $m \in \{\text{RGIR, NDSM}\}$. For each modality $m$, there are two branches: a Distinct branch ($m\_\text{Dist}$) that preserves modality-specific features and a Supplement branch ($m\_\text{Supp}$) that learns compensatory cues for an absent modality. During training, a randomly generated mask in $\{[\text{True, False}], [\text{False, True}]\}$ determines which Supplement branch is activated (\Cref{fig:inference_flow}). For instance, when NDSM is missing (mask $= [\text{True, False}]$), $\text{RGIR\_Supp}$ branch will be used in conjunction with the associated Distinct branch ($\text{RGIR\_Dist}$). Similarly, when RGIR is missing (mask $= [\text{False, True}]$), the $\text{NDSM\_Supp}$ branch will be paired with the $\text{NDSM\_Dist}$ branch. When both modalities are available, distinct branches including $\text{RGIR\_Dist}$ and $\text{NDSM\_Dist}$ are used. Hierarchical features from each branch $\{\boldsymbol{x}^m_i, \boldsymbol{x'}^{m}_i\}_{i=1}^4$ are progressively fused at each level, which are then passed into the Classwise Decoder.

\subsection{Disentanglement Learning with Knowledge Distillation (DLKD)}
\label{subsection:dlkd}
\textbf{\textit{Disentanglement Learning.}} Given an input image $\boldsymbol{I}^m$ and associated modality mask, each modality is decomposed into Distinct branch, preserving modality-specific features and Supplement branch, aiming to capture compensatory information that can act as a surrogate when another modality is absent. Formally, encoder features are produced as:
\begin{equation}
    \{\boldsymbol{x}^m_i\}_{i=1}^L = f_{\text{Dist}}^m(\boldsymbol{I}^m), \{\boldsymbol{x'}^m_i\}_{i=1}^L = f_{\text{Supp}}^m(\boldsymbol{I}^{m}),
\end{equation}
where $\boldsymbol{x}_i^m, \boldsymbol{x'}_i^m \in \mathbb{R}^{C_i \times W_i \times H_i}$ denote hierarchical feature maps at scale $i$, produced by Distinct and Supplement encoders $f_{\text{Dist}}^m$ and $f_{\text{Supp}}^m$, respectively. $L=4$ in our implementation. All encoders $f(\cdot)$ follow a convolutional UNet-style \cite{ronneberger2015u} design. To enforce separation between \textit{Distinct} and \textit{Supplement} features, we apply orthogonal constraint at every scale. The orthogonal loss $\mathcal{L}_{\text{orth}}$ is defined as the mean squared cosine similarity between L2-normalized features from Distinct and Supplement encoders.
\begin{equation}
    \mathcal{L}_{\text{orth}} = \sum_{i=1}^L\frac{1}{N}\sum^N_{j=1}\Bigg(\frac{\boldsymbol{x}_{ij}^m \cdot \boldsymbol{x'}_{ij}^m}{\|\boldsymbol{x}^m_{ij}\|_2 \|\boldsymbol{x'}_{ij}^m\|_2}\Bigg)^2,
\end{equation}
where $\boldsymbol{x}^m_{ij}, \boldsymbol{x'}^m_{ij} \in \mathbb{R}^{B \times D}$ are two sets of pooled feature embeddings generated from $f_{\text{Dist}}^m$ and $f_{\text{Supp}}^m$, respectively, for $j$-th sample at level $i$. $N$ is the number of samples in a batch. This formulation prevents Supplement branches from duplicating Distinct content, instead forcing them to encode non-overlapping information. To reduce the computational burden, original features are passed through attention pooling. In contrast to conventional disentanglement learning methods which separate modality-shared and modality-specific representations, our framework explicitly designs dedicated Supplement branches that are optimized to learn features capable of compensating for missing modality. This capability is enforced by knowledge distillation, where transferable privileged information from the full modality scenario is used as a supervisory guidance as discussed below.

\vspace{6px}
\noindent\textbf{\textit{Knowledge Distillation.}} To transfer such privileged knowledge from a full modality setting into incomplete modality settings, we adopt a hierarchical knowledge distillation strategy. The objective is to preserve as much transferable privileged information as possible in both intermediate feature representations and final logits when some modality is missing. For example, when NDSM is absent, structural cues should be inferred from the available spectral inputs, while in the absence of RGIR, spectral signatures should be compensated by structural information, ensuring that the fused representation remains close to the full-modality case. Our distillation loss is composed of two terms:
\begin{equation}
    \mathcal{L}_{\text{distill}} = \mathcal{L}_{\text{distill}}^{\text{feat}} + \mathcal{L}_{\text{distill}}^{\text{logit}}.
\end{equation}
Feature-level Distillation $\mathcal{L}_{\text{distill}}^{\text{feat}}$ is applied both to intermediate fused features across $L$ scales and to the penultimate representation produced by classwise decoder:
\begin{equation}
    \mathcal{L}_{\text{distill}}^{\text{feat}} = \sum_{i=1}^L\|\boldsymbol{F}_i^{\text{miss}}-\text{sg}(\boldsymbol{F}_i^{\text{full}})\|^2_2 + \|\boldsymbol{Z}^{\text{miss}} - \text{sg}(\boldsymbol{Z}^{\text{full}})\|^2_2,
\end{equation}
where $\boldsymbol{F}^{\text{miss}}_{i}, \boldsymbol{F}^{\text{full}}_i$ denote fused features of missing modality scenario and full modality setting at level $i$, respectively. $\text{sg}(\cdot)$ indicates stop-gradient to freeze teacher (full modality) representations. $\boldsymbol{Z}$ is the penultimate feature map $\boldsymbol{Z} \in \mathbb{R}^{C \times H \times W}$. For computational efficiency, fused features are attention pooled before MSE calculation except for penultimate features. On the other hand, Logit-level Distillation $\mathcal{L}_{\text{distill}}^{\text{logit}}$ encourages consistency at the decision boundary by minimizing the Kullback–Leibler divergence between temperature-scaled logits from the teacher (full modality) and student (missing modality).
\begin{equation}
    \mathcal{L}_{\text{distill}}^{\text{logit}} = T^2\times \text{KL}\Bigg(\text{soft}\Big(\frac{\text{sg}(\boldsymbol{z}^{\text{full}})}{T}\Big) \; \Bigg\| \; \text{log\_soft}\Big(\frac{\boldsymbol{z}^{\text{miss}}}{T}\Big) \Bigg),
\end{equation}
where $\boldsymbol{z}$ are logits and $T=2$ is the distillation temperature. $\text{soft}(\cdot), \text{log\_soft}(\cdot)$ are softmax and log softmax functions, respectively. Our final DLKD loss term is defined as follow:
\begin{equation}
    \mathcal{L}_{\text{DLKD}} = \mathcal{L}_{\text{orth}} + \mathcal{L}_{\text{distill}}.
\end{equation}
This combined formulation ensures that Supplement features (i) remain non-overlapping from Distinct cues and (ii) inherit transferable semantics from full-modality setting, thereby preserving predictive performance under missing modality scenarios.

\subsection{Hierarchical Hybrid Fusion (HF)}
\label{subsection:hf}
This structure is designed to progressively integrate multimodal features across pyramid levels. At the lower levels, convolutional fusion captures local information, while at the last level, transformer-based fusion provides global context through long-range dependencies. Let $\boldsymbol{a}_i, \boldsymbol{b}_i\in \mathbb{R}^{C_i \times H_i \times W_i}$ denote features from two active branches depending on scenario mask at each pyramid level $i \in \{1, 2, 3\}$. These features are concatenated $\text{Concat}(\cdot)$ and fused through a convolutional-based fusion block $\phi_i(\cdot)$:
\begin{equation}
    \boldsymbol{F}_i = \phi_i(\text{Concat}(\boldsymbol{a}_i, \boldsymbol{b}_i)).
\end{equation}
To preserve hierarchical consistency, residual connections are used to link adjacent levels. At the last level of the encoder, a transformer-based fusion block $\mathcal{T}(\cdot)$ operates on concatenated sequence with positional embeddings.
\begin{equation}
    \boldsymbol{F}_L = \mathcal{T}(\text{Concat}(\boldsymbol{t}^*, \boldsymbol{a}_L, \boldsymbol{b}_L), \boldsymbol{F}_{L-1}),
\end{equation}
where $\boldsymbol{a}_L, \boldsymbol{b}_L$ denotes features at the last layer $L$ from separate activated encoder branches. The fused features from the previous level $\boldsymbol{F}_{L-1}$ are pooled to reduce computation cost. A learnable fusion token $\boldsymbol{t}^* \in \mathbb{R}^{1\times d}$ is introduced to jointly attend across modalities. It will serve as a compact representation of the last layer in Feature-level Distillation.

\subsection{Classwise Feature Learning Module (CFLM)}
\label{subsection:cflm}
The significance of each modality contribution varies across classes. Especially, under missing modality scenario, the model must adaptively rely on different information to generate a reliable prediction for each class. Therefore, we introduce the Classwise Feature Learning Module (CFLM), consisting of two parts: Classwise Attention and Classwise Decoder (\Cref{fig:model_schema}B). Our design enables the network to emphasize discriminative spatial regions per class and refine predictions at multiple scales. Classwise Attention component receives the fused features of the last encoder layer $\boldsymbol{F}_L$ and assigns each semantic class $k \in \{1,...,K\}$ a learnable query vector $\boldsymbol{q}_k \in \mathbb{R}^C$. For each class, an attention score is computed as:
\begin{equation}
    \boldsymbol{\alpha}_k= \text{soft}\Bigg(\frac{\boldsymbol{q}_k \boldsymbol{F}_{L}}{\sqrt{C}}\Bigg).
\end{equation}
Raw attention scores are put through Softmax $\text{soft}(\cdot)$. The attended class-specific feature maps are obtained by:
\begin{equation}
    \boldsymbol{M}_k = \boldsymbol{\alpha}_k\boldsymbol{F}_{L}, \boldsymbol{M}_k \in \mathbb{R}^{C\times \frac{H}{16} \times \frac{W}{16}}.
\end{equation}
Each classwise attention map (CWAM) $\boldsymbol{M}_k$ is passed to a hierarchical classwise decoder together with fused features from earlier scales $\{\boldsymbol{F}_i\}_{i=1}^{L-1}$. The decoder progressively upsamples $\text{Up}(\cdot)$ and refines predictions via convolution-based block $\phi(\cdot)$ for each class.
\begin{equation}
\boldsymbol{D}_k^{\,l} =
\begin{cases}
\phi_k^{\,l}\!\left([\;\boldsymbol{F}_l,\; \mathrm{Up}(\boldsymbol{D}_k^{\,l+1})\;]\right), & l \in \{1,2,3\}, \\[6pt]
\phi_k^{\,4}\!\left(\boldsymbol{M}_k\right), & l = 4.
\end{cases}
\end{equation}
To form the penultimate feature map $\boldsymbol{Z}$ for knowledge distillation and for generating final semantic segmentation, decoded outputs from all classes are concatenated $\text{Concat}(\cdot)$ and aggregated via convolution $\phi(\cdot)$:
\begin{equation}
    \boldsymbol{Z} = \phi(\text{Concat}(\|^K_{k=1}\boldsymbol{D}_k^1)), \boldsymbol{Z} \in \mathbb{R}^{C \times H \times W}.
\end{equation}
In addition, at each level, the decoder produces auxiliary segmentation map using convolution and upsampling to enhance multiscale learning. The auxiliary loss is defined as:
\begin{equation}
    \mathcal{L}_{\text{aux}}= \sum_{l=1}^{L}\mathcal{L}_{\text{seg}}(\boldsymbol{\hat{Y}}^l, \boldsymbol{Y}),
\end{equation}
where $\boldsymbol{\hat{Y}}^l$ is the prediction at level $l$ and $\boldsymbol{Y}$ is the ground-truth semantic segmentation map. To relieve severe class imbalance issue in remote sensing datasets, the $\mathcal{L}_{\text{seg}}$ loss is the combination of weighted Cross Entropy loss and Dice loss. The final segmentation map $\boldsymbol{\hat{Y}}_{\text{seg}}$ are formed by applying $1\times1$ convolution to the penultimate feature map $\boldsymbol{Z}$.

\begin{figure}
  \centering
  \includegraphics[width=\linewidth]{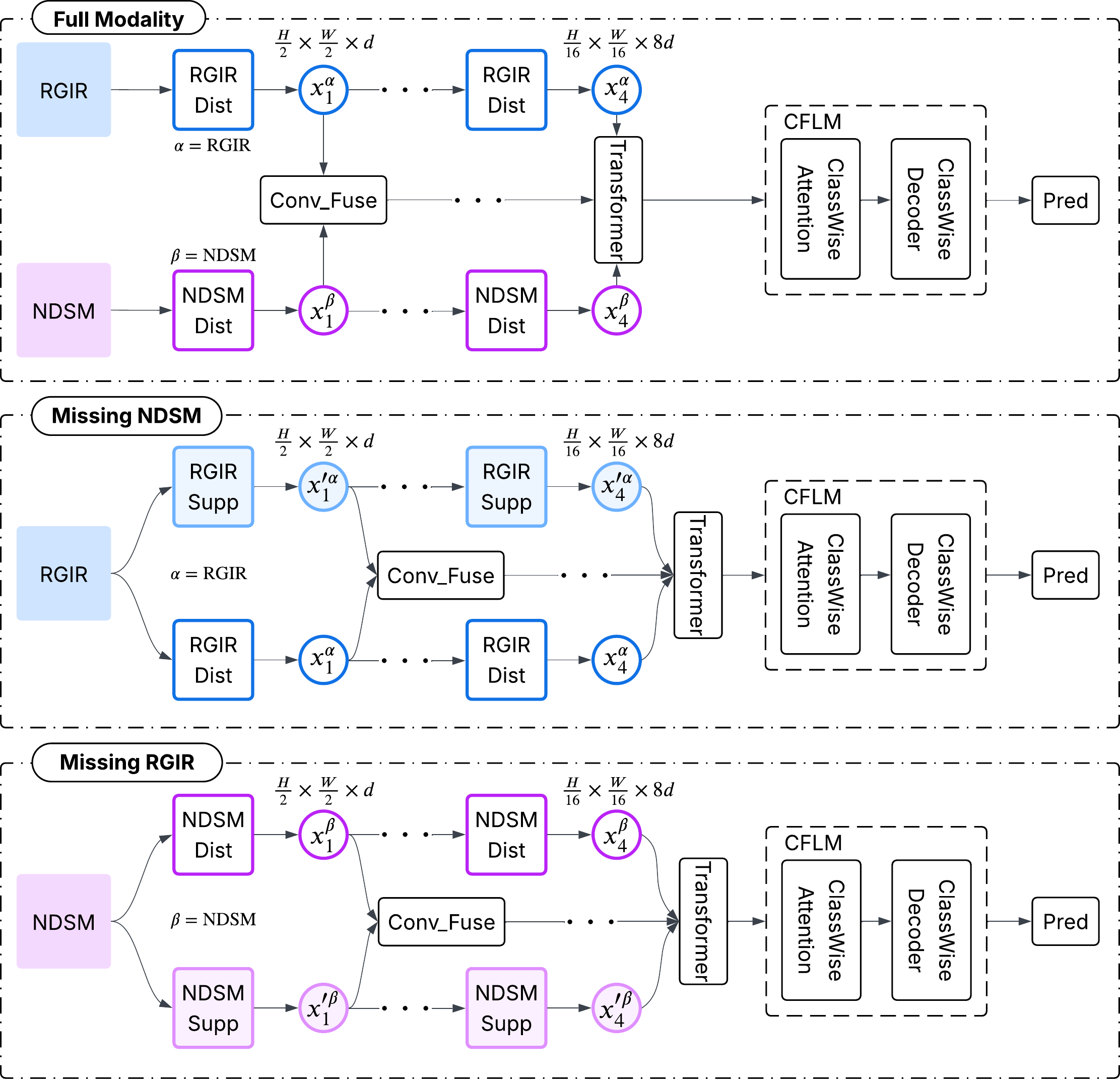}
  \caption{Model flow based on scenario at inference time}
  \label{fig:inference_flow}
\end{figure}

\vspace{3px}
The overall training loss function of our framework is the combination of minimizing all objectives discussed above:
\begin{equation}
    \mathcal{L} = \mathcal{L}_{\text{seg}}(\boldsymbol{\hat{Y}}_{\text{seg}}^{\text{full}}, \boldsymbol{Y}) + \mathcal{L}_{\text{seg}}(\boldsymbol{\hat{Y}}_{\text{seg}}^{\text{miss}}, \boldsymbol{Y}) + \mathcal{L}_{\text{DLKD}} + \sum\mathcal{L}_{\text{uni}}^m + \mathcal{L}_{\text{aux}},
\end{equation}
where $\mathcal{L}_{\text{seg}}(\boldsymbol{\hat{Y}}_{\text{seg}}^{\text{full}}, \boldsymbol{Y})$ corresponds to the segmentation loss under the full modality scenario. The term $\mathcal{L}_{\text{seg}}(\boldsymbol{\hat{Y}}_{\text{seg}}^{\text{miss}}, \boldsymbol{Y})$ denotes segmentation loss of missing modality setting, where depending on scenario mask, a particular pair of Distinct and Supplement branches is activated (as shown in \Cref{fig:inference_flow}). To further encourage the model to learn discriminative features, unimodal segmentation loss $\mathcal{L}_{\text{uni}}^m$ is applied to each individual modal $m \in \{\text{RGIR, NDSM}\}$.

%% file: sec/4_experiment.tex
\section{Experiments}
\subsection{Experimental Setting}
Experiments are conducted on two popular RS datasets that offer fine-grained label for semantic segmentation: \textbf{Vaihingen \cite{Vaihingen}} and \textbf{Potsdam \cite{Potsdam}}. Both datasets are from the International Society for Photogrammetry and Remote Sensing (ISPRS) and have two modalities: RGIR and NDSM. Vaihingen includes 33 large tiles of different sizes. On the other hand, Potsdam dataset contains 38 patches of the same size. Both datasets suffer from severe class imbalance, with the `car' class accounting for only about $1\%$.

\vspace{3px}
\noindent Samples of size $512\times512$ are obtained using dedicated training, validation and test tiles specified on the ISPRS portal. To increase diversity, random augmentations are applied. All models reported in this paper were trained on an NVIDIA GeForce RTX 3090. Class-wise F1 score, mean F1 score (mF1), and mean Intersection over Union (mIoU) are evaluation metrics used for performance analysis.

\vspace{3px}
Our method \textbf{DIS2} is compared against a diverse set of SOTA models, covering different methodological families. GEMMNet~\cite{kieu2025gemmnet} exemplifies the generative paradigm, with design choices tailored to the challenges of RS data. Mmformer~\cite{zhang2022mmformer} is a powerful transformer-based architecture, originally proposed for medical imaging, that implicitly learns cross-modal dependencies. ShaSpec~\cite{wang2023multi} is a representative work using disentanglement learning in medical imaging. SimSiam~\cite{chen2021exploring} and our adapted SimSiamTransfer baseline, represent knowledge distillation strategies through contrastive learning without the need for negative samples. 

\subsection{Result Analysis}
%This section provides a comprehensive evaluation of the robustness towards missing modality of our proposed framework \textbf{DIS2} in remote sensing semantic segmentation. 
Results are presented across three scenarios on two datasets: (1) Full Modality, (2) Missing RGIR (only NDSM available), and (3) Missing NDSM (only RGIR available).

\subsubsection{Quantitative Analysis:}
\Cref{tab:vaihingen_potsdam} reports class-wise F1 scores, mF1, and mIoU on the Vaihingen and Potsdam benchmarks under three evaluation settings mentioned above.
\begin{table*}[t]
  \centering
  \caption{\textbf{Results on Vaihingen and Potsdam datasets.} Class-wise F1 scores are reported along with the performance gap between our model and competing approaches ($\Delta$). Models mmformer \cite{zhang2022mmformer}, shaspec \cite{wang2023multi}, SimSiam \cite{chen2021exploring} and SimSiamTransfer are our own implementations, adapted from SOTA methods in other domains to the RS data. Results for GEMMNet \cite{kieu2025gemmnet} are taken directly from the original paper.}
  \label{tab:vaihingen_potsdam}
  \includegraphics[width=\textwidth]{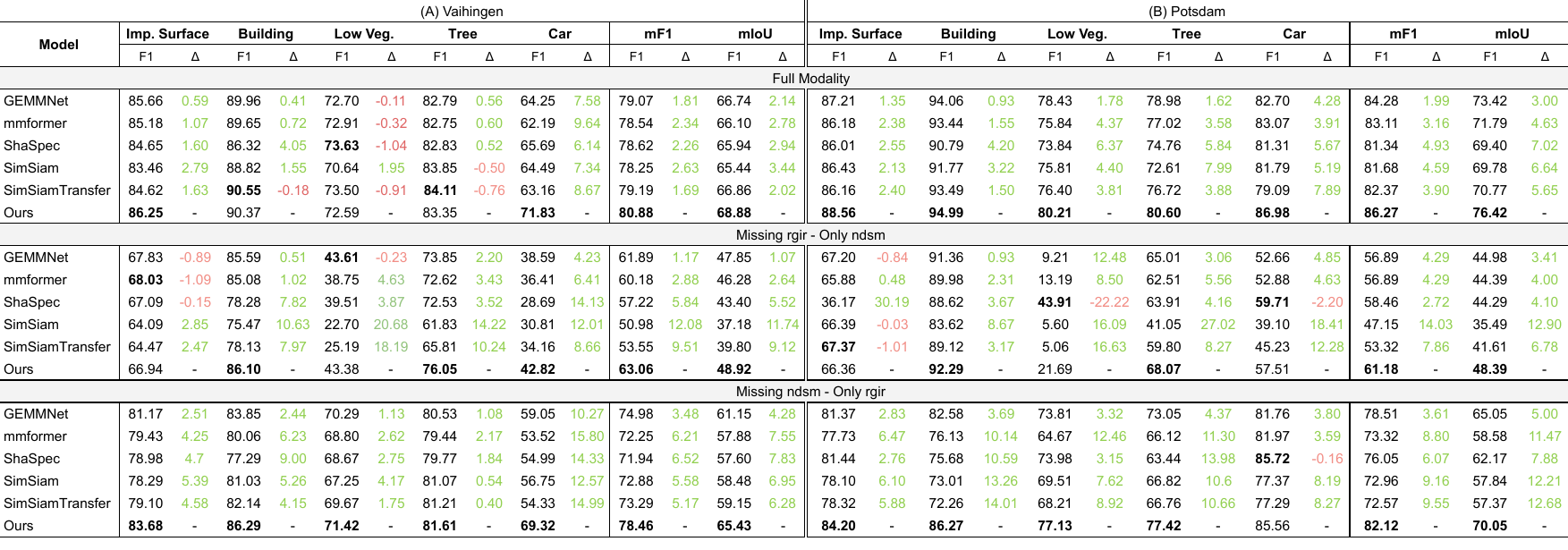}
\end{table*}

\vspace{3px}
\noindent\textbf{\textit{Full Modality.}} When both modalities are present, our model achieves the highest performance across both datasets. The proposed method shows consistent improvements of $+2.14$ and $+3.00$ points mIoU over the strongest competitor - GEMMNet. Our model surpasses the representative disentanglement learning method - ShaSpec - by $+7.02$ points mIoU on Potsdam. Notably, the largest gains are observed in the challenging `car' class,  with F1 score improvements ranging from $+6.14$ to $+9.64$ points on Vaihingen and from $+3.91$ to $+7.89$ points on Potsdam. This observation confirms that our method effectively mitigates scale variation and class imbalance.

\vspace{3px}
\noindent\textbf{\textit{Missing RGIR.}} This is the most challenging scenario because RGIR information is richer than NDSM. All other competitive methods show significant performance degradation. Our method outperforms the next best competitor by $+1.17$ mF1 and $+1.07$ mIoU on Vaihingen, and by $+2.72$ mF1 and $+3.41$ mIoU on Potsdam. Importantly, the vulnerable `car' class, greatly benefits from our design, achieving F1 score boost from $+4.23$ to $+14.13$ on Vaihingen and from $+4.63$ to $+18.41$ on Potsdam. These results highlight the effectiveness of the NDSM\_Supplement branch, guided by distillation, in compensating for lost spectral cues.

\vspace{3px}
\noindent\textbf{\textit{Missing NDSM.}} Under this setting, the performance drop is less pronounced. Our model still demonstrates clear superiority, reaching $78.46$ mF1 on Vaihingen, compared to less than $75.00$ mF1 for all competing methods, and $65.43$ mIoU, while others remain around $61.00$ or lower. Similarly, on Potsdam, our approach delivers consistent gains, with mF1 improvements varying between $+3.61$ to $+9.55$ and mIoU enhancement from $+5.00$ to $+12.68$ over existing methods. The car class on Vaihingen shows significant gains, outperforming the second-best model (GEMMNet) by $+10.27$ mF1 and achieving a maximum improvement of $+15.80$ mF1 compared to mmformer.

\subsubsection{Qualitative Analysis:}
\begin{figure*}
  \centering
  \includegraphics[width=0.85\textwidth]{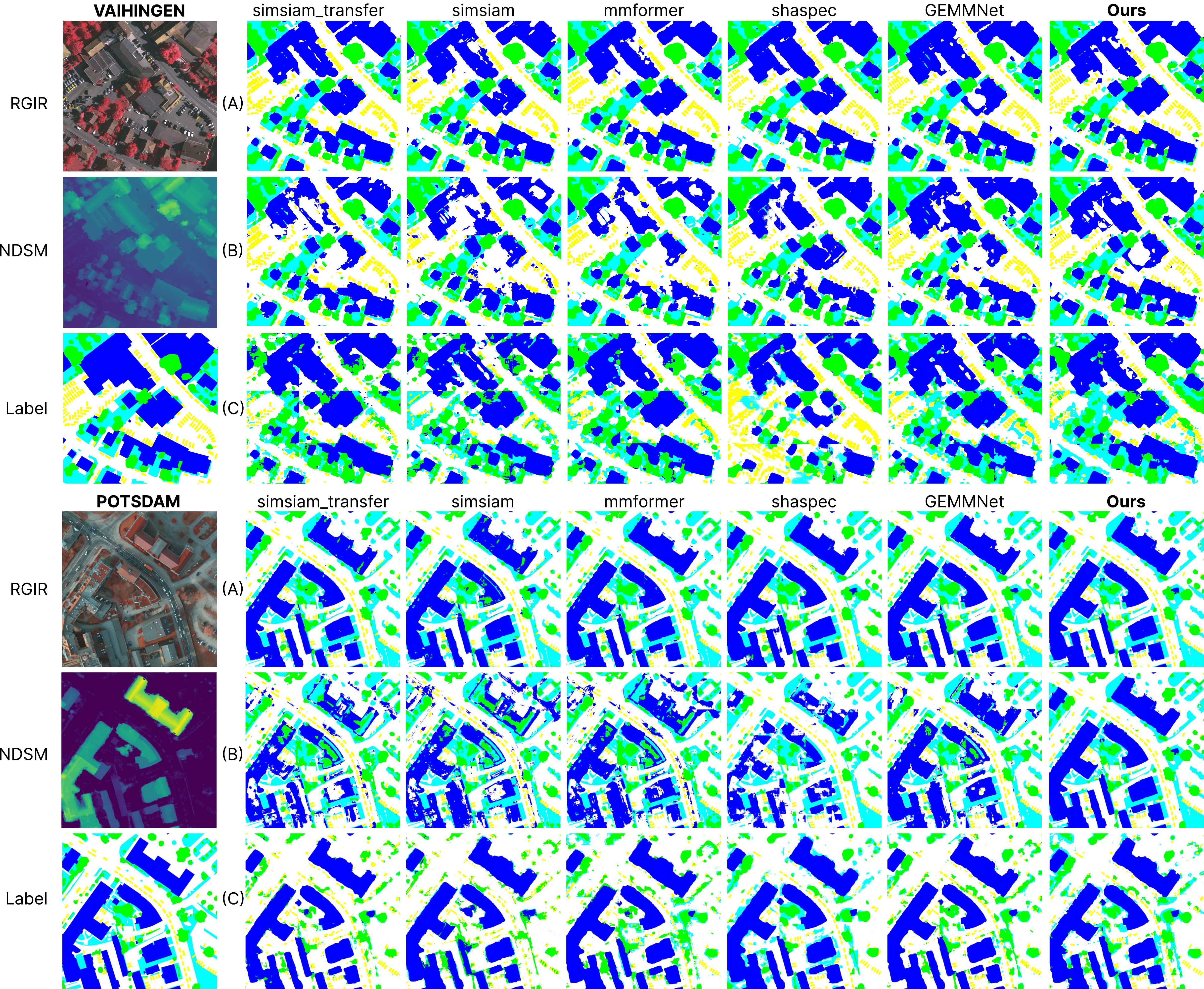}
  \caption{Qualitative results on the Vaihingen and Potsdam datasets under different modality settings. Our method remains robust under missing-modality scenarios, producing segmentation maps that closely match ground truth, with more intact predictions for buildings, roads, and vegetation, and sharper object boundaries for cars. (A) Full Modality, (B) Missing NDSM, and (C) Missing RGIR.}
  \label{fig:qualitative}
\end{figure*}

\Cref{fig:qualitative} presents qualitative comparisons on the Vaihingen and Potsdam datasets under different modality settings. We visualize predictions from competitive methods (GEMMNet, mmformer, ShaSpec, SimSiam, SimSiamTransfer) alongside our method. Ground-truth labels and modality inputs (RGIR, NDSM) are also provided for reference.

On Vaihingen, our method produces segmentation maps that align more closely with ground truth, particularly in structurally complex regions such as building and narrow roads. \textit{Our model is better at preserving clear object boundaries in `car' class.} On Potsdam, our method again yields the most visually coherent results. \textit{Other competitive methods struggle with shadows and occlusions, leading to fragmented predictions of building and roads.} Especially under the missing NDSM setting, existing SOTA approaches frequently confuse buildings with tree and low veg producing unrealistic fragmented semantic segmentation map. \textit{Our method better delineates object boundaries and maintains intact, consistent predictions across large-scale structures like building.} The superiority of our model is most evident under missing modality conditions. When NDSM is absent, other competitors relying heavily on height cues suffer from collapsed building predictions. Our RGIR\_Supplement branch, guided by knowledge distillation, capable of restoring some structural information. When RGIR is missing, noisy vegetation boundaries are produced by other methods; however, our \textbf{DIS2} retains sharper contours thanks to compensating cues from NDSM\_Supplement branch.

\subsubsection{Ablation Study:}
\noindent\textbf{\textit{Effectiveness of Different Components.}} To assess the contribution of each component in our framework, we conducted ablation experiments on Vaihingen (\Cref{tab:ablation}) by incrementally enabling Hierarchical Hybrid Fusion (HF), Classwise Feature Learning Module (CW), and Disentanglement Learning with Knowledge Distillation (DLKD). Using HF alone provides a strong baseline in full modality mode. Adding CW and DLKD separately into HF-enabled models also yields similar performance. When all three components are combined, our model achieves the best results. \textit{Crucially, the synergy of CW and DLKD when integrated with HF is clearly demonstrated under missing modality settings, which is the main objective of our study.} 

When missing RGIR, adding CW and DLKD separately improves recognition of rare `car' class to $40.72$ and $39.21$ from $30.02$. Similarly, `Tree' class improvements are also noticeable from $68.36$ to over $74$ for both modules. The best performance is obtained across all classes when three components are integrated. Under missing NDSM condition, addition of CW and DLKD separately improves prediction by about $+4.0$ points to over $67$. Our fully integrated model again delivers the highest performance on `car' class, mF1 and mIoU. Across all scenarios, our complete model (HF + CW + DLKD) achieves the strongest results, showing that \textit{(i) HF establishes a solid multimodal baseline, (ii) CW enhances recognition of small and underrepresented classes, and (iii) DLKD transfers privileged knowledge from full modality setting to improve robustness under missing-modality conditions}.
\begin{table}
  \caption{Ablation results on Vaihingen dataset (F1 score).}
  \label{tab:ablation}
  \includegraphics[width=\linewidth]{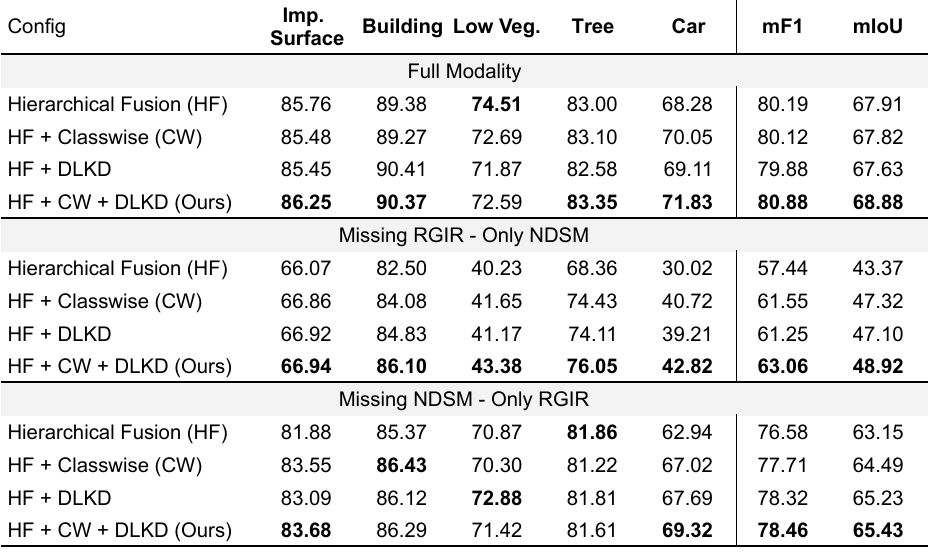}
\end{table}

\begin{figure}
  \centering
  \includegraphics[width=\linewidth]{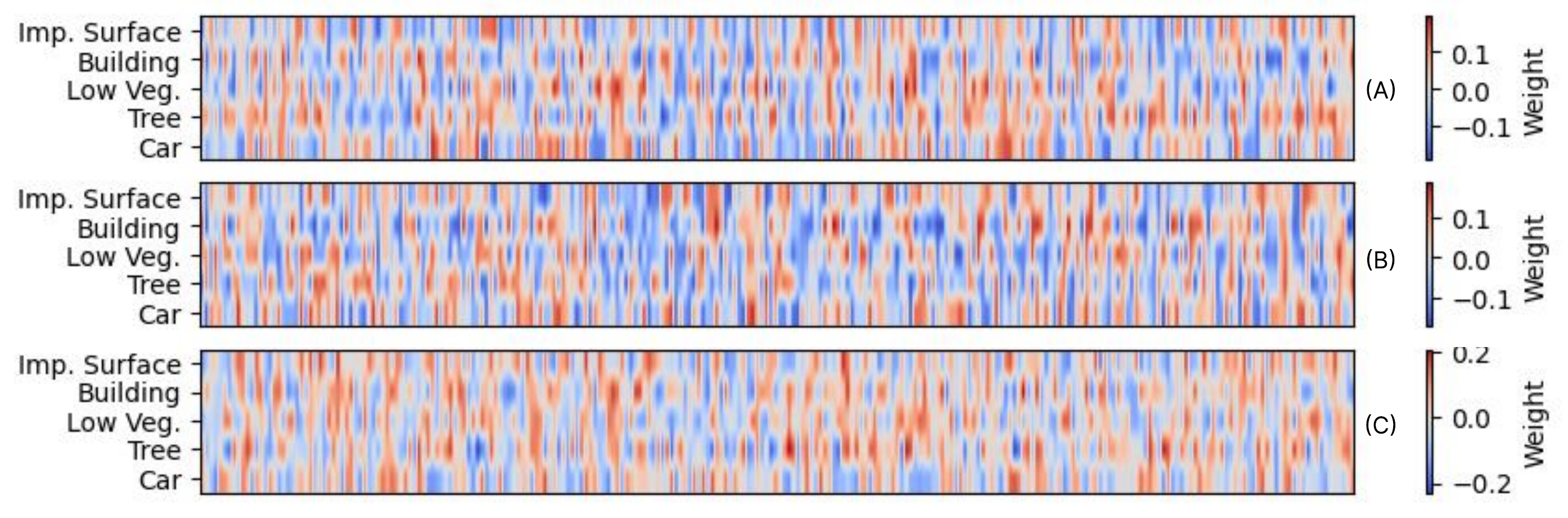}
  \caption{Classwise Learnable Query Heatmap}
  \label{fig:query_heatmap}
\end{figure}

\begin{figure}
  \centering
  \includegraphics[width=\linewidth]{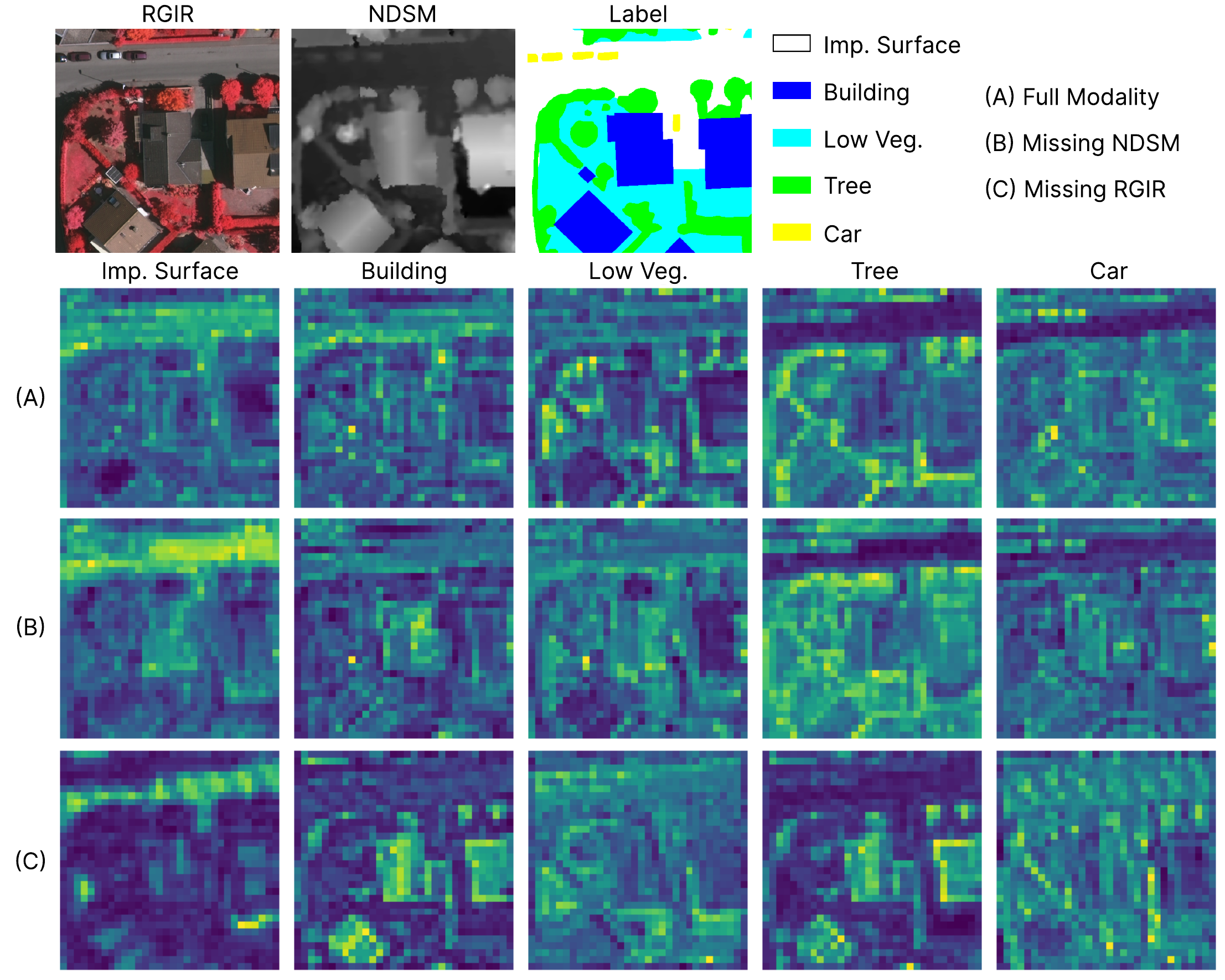}
  \caption{Visualization of classwise attention maps under different modality settings on Vaihingen dataset. Focus adaptively shifts depending on targeting class and modality availability.}
  \label{fig:CWAM}
\end{figure}

\noindent\textbf{\textit{Visualization of Classwise Attention.}} To further analyze the role of CFLM, we visualize its behaviour through classwise attention maps and query heatmaps. As in \Cref{fig:CWAM}, \textit{attention is distributed discriminatively} across different semantic categories. Even within the same class, the activation patterns vary depending on scenarios. Moreover, \Cref{fig:query_heatmap} reveals distinguishable query patterns across classes and scenarios, indicating that the model `looks' for different cues through query to selectively focus on class-relevant details producing adaptive classwise attention maps. These results demonstrate that our design provides an effective mechanism for each semantic class to seek compensating missing information through selective attention. 

\begin{figure}
  \centering
  \includegraphics[width=0.85\linewidth]{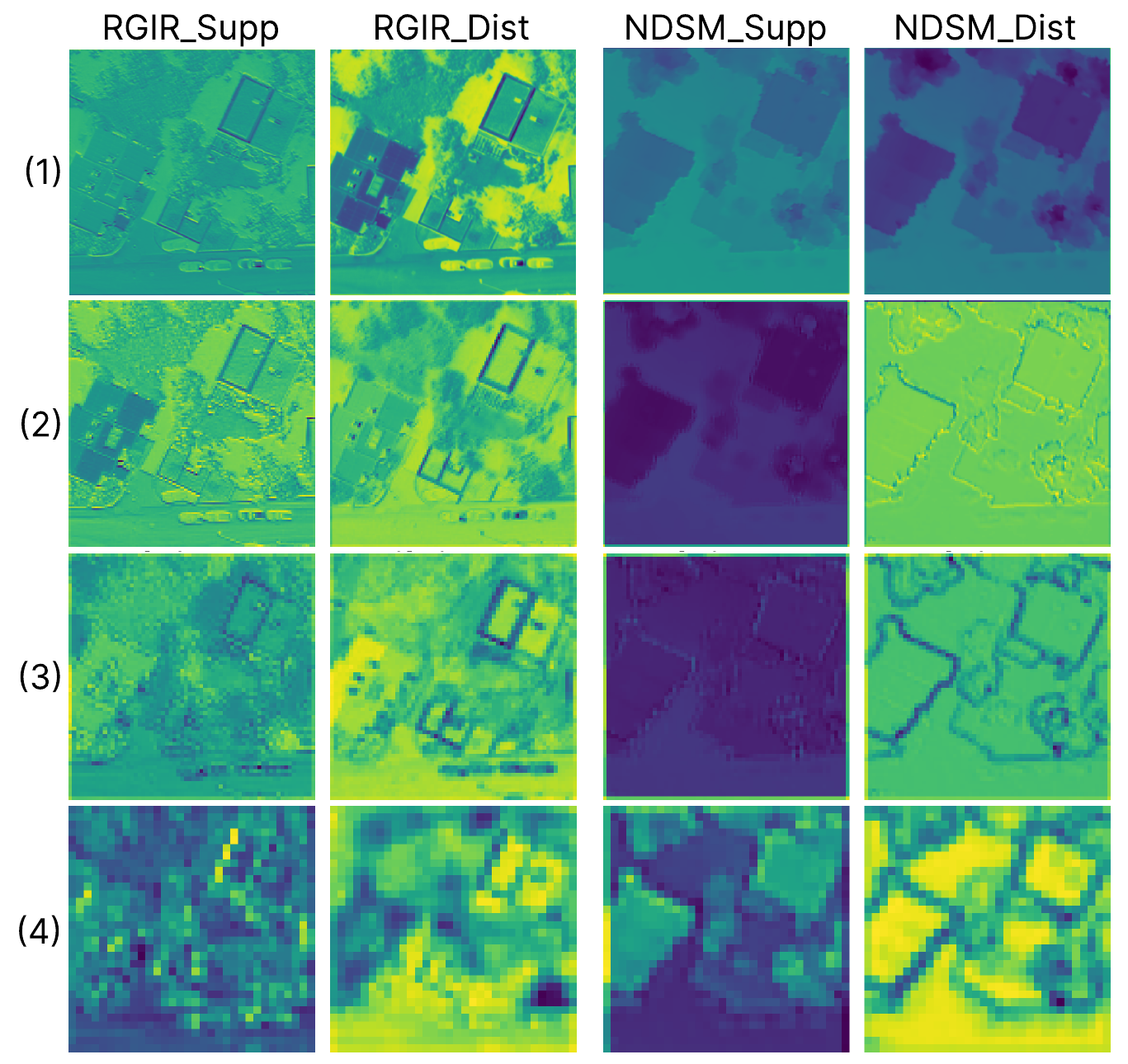}
  \caption{Intermediate feature activation maps at 4 pyramid levels}
  \label{fig:intermediate_feature}
\end{figure}

\vspace{3px}
\noindent\textbf{\textit{Visualization of DLKD.}}
The impact of DLKD is examined by analyzing intermediate feature maps from the Distinct and Supplement branches. Distances between penultimate features across scenarios are also compared for our \textbf{DIS2} and competing models. \Cref{fig:intermediate_feature} shows outputs of different branches under various modality cases. The Supplement branches \textit{consistently attend to different details} compared to the Distinct ones, highlighting their role in handling missing modalities. Their activation patterns increase the richness and coverage of representations to the Distinct branches, effectively compensating for the absence of the other modality. 

\Cref{tab:distance} reports the average L2 distances between penultimate features across scenarios. \textit{Our method achieves the lowest feature discrepancies in all cases, suggesting stronger alignment between missing modality prediction and the ideal full modality setting.} The results verify that our DLKD strategy significantly narrows the representational gap between scenarios, yielding more stable and consistent outputs. Together, the intermediate feature visualizations and distance measurements confirm that DLKD \textit{achieves two key objectives: (i) enforcing disentanglement so that Supplement branches learn compensating rather than redundant information, and (ii) leveraging knowledge distillation to align missing modality predictions with full modality semantics.}
\begin{table}
  \caption{Average L2 distance between penultimate features of different scenarios $(\times10^{-4})$. A smaller distance indicates better alignment between prediction spaces.}
  \label{tab:distance}
  \includegraphics[width=\linewidth]{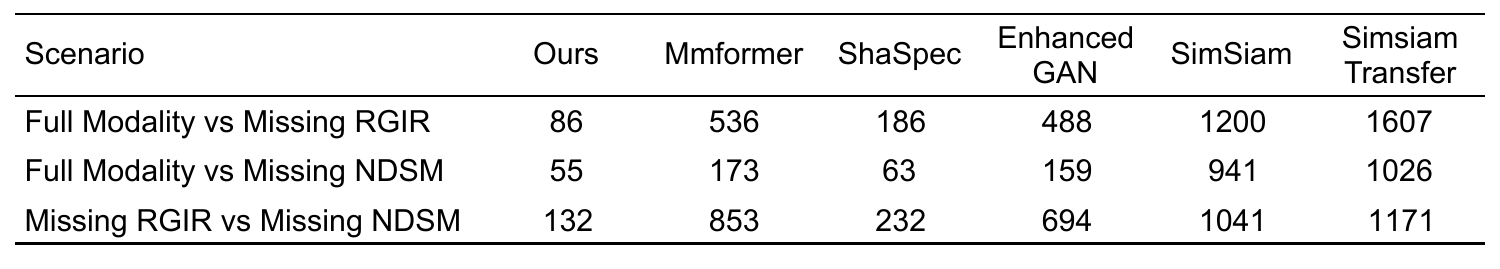}
\end{table}

%% file: sec/5_conclusion.tex
\section{Conclusion}
In this paper, we addressed the multifaceted challenge of missing modalities in RS semantic segmentation. Existing paradigms are ill-equipped for this domain due to unique data characteristics of heterogeneous signals, huge scale variation and severe class imbalance. To overcome these limitations, we introduced DIS2, a novel framework built on a holistic approach of active, guided feature compensation. Our work is grounded in three pillars: principled information compensation, class-specific modeling, and multi-scale learning. We presented a reformulated synergy of disentanglement and knowledge distillation, DLKD, which transforms a passive feature decomposition process into a targeted reconstruction task, where a dedicated pathway learns to act as an expert surrogate for the missing data. Moreover, through our Classwise Feature Learning Module (CFLM), we ensure that the model is sensitive to the varying importance of modalities for each semantic class under different scenarios. The entire process of DLKD and CFLM is supported by a Hierarchical Hybrid Fusion (HF) backbone, ensuring robustness across all scales. DIS2 consistently and significantly outperformed a suite of state-of-the-art methods in full-modality, and more importantly, in various missing modality scenarios. The results confirmed that our holistic, RS-aware approach provides a robust solution.

\vspace{-2mm}
\section*{Acknowledgment}
This work partly supported by Shield AI - one of the global leaders in developing AI pilots for defence and civilian applications.
\vspace{-1.5mm}

%\enlargethispage{2\baselineskip}